\documentclass[acmsmall,screen,nonacm]{acmart}
\usepackage{subcaption}
\usepackage{pdflscape}
\usepackage{multirow}
\usepackage[normalem]{ulem}

\usepackage{tikz}
\usetikzlibrary{shapes}

\newcommand{\ie}{{\it i.e. }}
\newcommand{\eg}{{\it e.g. }}

\newcommand{\linf}{{$\ell_{\infty}$}}
\newcommand{\ltwo}{{$\ell_{2}$}}

\newcommand{\lzero}{{$\ell_{0}$}}

\definecolor{darkgreen}{RGB}{0, 100, 0}
\definecolor{darkblue}{RGB}{0, 0, 128}

\newcommand{\checkdef}{\textcolor{darkgreen}{\checkmark}}
\newcommand{\defense}[1]{\textcolor{darkgreen}{#1}}
\newcommand{\checkatt}{\textcolor{darkblue}{\checkmark}}
\newcommand{\attack}[1]{\textcolor{darkblue}{#1}}

\newcommand{\xadv}{\ensuremath{x_{adv}}}
\newcommand{\normp}[2]{\|#2\|_{#1}}

\DeclareMathOperator*{\EE}{\mathbb{E}}
\newcommand{\D}{\ensuremath{\mathcal{D}}}
\newcommand{\normal}{\ensuremath{\mathcal{N}}}

\newcommand{\PGD}{{\sc pgd}}
\newcommand{\IFGSM}{{\sc ifgsm}}
\newcommand{\FGSM}{{\sc fgsm}}
\newcommand{\LBFGS}{{\sc l-bfgs}}
\newcommand{\DKNN}{{\sc dknn}}
\newcommand{\JSMA}{{\sc jsma}}
\newcommand{\deepfool}{{\sc deepfool}}
\newcommand{\LMT}{{\sc lmt}}
\newcommand{\GDA}{{\sc gda}}
\newcommand{\YOPO}{{\sc yopo}}
\newcommand{\CW}{{\sc c}\&{\sc w}}
\newcommand{\ParsevalTraining}{{\sc parseval training}}

\newcommand{\reluplex}{{\sc reluplex}}
\newcommand{\reluval}{{\sc reluval}}
\newcommand{\marabou}{{\sc marabou}}
\newcommand{\aiai}{{\sc ai$^2$}}
\newcommand{\DLV}{{\sc dlv}}
\newcommand{\refinezono}{{\sc refinezono}}
\newcommand{\lirpa}{{\sc lirpa}}

\newcommand{\LIME}{{\sc lime}}
\newcommand{\anchors}{{\sc anchors}}
\newcommand{\SHAP}{{\sc shap}}
\newcommand{\NPSEM}{{\sc npsem}}
\newcommand{\LORE}{{\sc lore}}
\newcommand{\GVE}{{\sc gve}}
\newcommand{\maskingmodel}{{\sc masking model}}
\newcommand{\MPSM}{{\sc mpsm}}
\newcommand{\deconvnet}{{\sc deconvnet}}
\newcommand{\deeplift}{{\sc deeplift}}

\newcommand{\agnostic}{{\sc all}}
\newcommand{\NN}{
\scalebox{.45}{
\begin{tikzpicture}
    \tikzset{weights/.style={-}}
    \node[draw,circle,scale=0.7] (i1)at(0,0){};
    \node[draw,circle,scale=0.7] (i2)at(0,-0.4){};
    \node[draw,circle,scale=0.7] (i3)at(0,-0.8){};
    \node[draw,circle,scale=0.7] (h11)at(0.4,-0.2){};
    \node[draw,circle,scale=0.7] (h12)at(0.4,-0.6){};
    \node[draw,circle,scale=0.7] (h21)at(0.8,-0.4){};
    \draw[weights] (-0.3,0)--(i1);
    \draw[weights] (-0.3,-0.4)--(i2);
    \draw[weights] (-0.3,-0.8)--(i3);
    \draw[weights] (i1)--(h11);
    \draw[weights] (i1)--(h12);
    \draw[weights] (i2)--(h11);
    \draw[weights] (i2)--(h12);
    \draw[weights] (i3)--(h11);
    \draw[weights] (i3)--(h12);
    \draw[weights] (h11)--(h21);
    \draw[weights] (h12)--(h21);
    \draw[weights] (h21)--(1.1,-0.4);
\end{tikzpicture}%
}
}

\AtBeginDocument{%
  \providecommand\BibTeX{{%
    \normalfont B\kern-0.5em{\scshape i\kern-0.25em b}\kern-0.8em\TeX}}}

\title{Certification of embedded systems based on Machine Learning: A survey}

\author{Guillaume Vidot}
\email{eric-guillaume.vidot@airbus.com}
\email{eric.vidot@irit.fr}
\affiliation{%
  \institution{Airbus Opération S.A.S}
  \city{Toulouse}
  \country{France}
}
\affiliation{%
  \institution{University of Toulouse, IRIT}
  \city{Toulouse}
  \country{France}}

\author{Christophe Gabreau}
\affiliation{%
  \institution{Airbus Opération S.A.S}
  \city{Toulouse}
  \country{France}}
\email{christophe.gabreau@airbus.com}

\author{Ileana Ober}
\affiliation{%
  \institution{University of Toulouse, IRIT}
  \city{Toulouse}
  \country{France}}
\email{ileana.ober@irit.fr}

\author{Iulian Ober}
\affiliation{%
  \institution{University of Toulouse, IRIT}
  \city{Toulouse}
  \country{France}}
\email{iulian.ober@irit.fr}

\renewcommand{\shortauthors}{Vidot, et al.}

\begin{document}

\begin{abstract}
  Advances in machine learning (ML) open the way to innovating functions in the avionic domain, such as navigation/surveillance assistance (e.g. vision-based navigation, obstacle sensing, virtual sensing), speech-to-text applications, autonomous flight, predictive maintenance or cockpit assistance. Current certification standards and practices, which were defined and refined decades over decades with classical programming in mind, do not however support this new development paradigm. This article provides an overview of the main challenges raised by the use ML in the demonstration of compliance with regulation requirements, and a survey of literature relevant to these challenges, with particular focus on the issues of robustness and explainability of ML results.
\end{abstract}

\maketitle

\section{Introduction}\label{sec:intro}

Safety critical systems, and avionics in particular, represent an application field unenthusiastic in applying new software development methods \cite{Krodel2008}, as shown by the fact that some aspects of the object-oriented programming, such as polymorphism and dynamic biding, never made their way to safety critical systems, mostly because the inconvenient balance between added value and increased development and validation costs.
Nevertheless, the recent advances in machine learning triggered genuine interest, as machine learning offer promising preliminary results and open the way to a wide range of new functions for avionics systems, for instance in the area of autonomous flying. In this paper we investigate on  how existing certification and regulation techniques, can (or cannot) handle software development that includes parts obtained by machine learning.

Nowadays a large aircraft cockpit offers many avionic complex functions: flight controls, navigation, surveillance, communications, displays... Their design has required a top down iterative approach from aircraft level downward, thus the functions are performed by systems of systems, with each system decomposed into subsystems that may contain a collection of software and hardware items.
Therefore, any avionic development considers 3 levels of engineering: {\it (i)} Function, {\it (ii)} System/Subsystem and {\it (iii)} Item.
The development process of each engineering level relies on several decades of experience and good practices that keep on being adapted today.
These methods have been standardized through
EUROCAE/SAE standards for system development (ED-79A/ARP4754A) and EUROCAE/RTCA standards for software items (ED-12C/DO-178C and supplements)
and hardware items (ED-80/DO-254). They are recognized as applicable means of compliance with regulation requirements by worldwide Avionic Authorities (AAs).
In this context, this survey actually reduces the domain of study to the item level and more precisely to the software item.
Note that at the item level, the correct terminology for the demonstration of conformity to standards is ``qualification''.

\begin{figure}[b!]
    \centering
    \includegraphics[width=\textwidth]{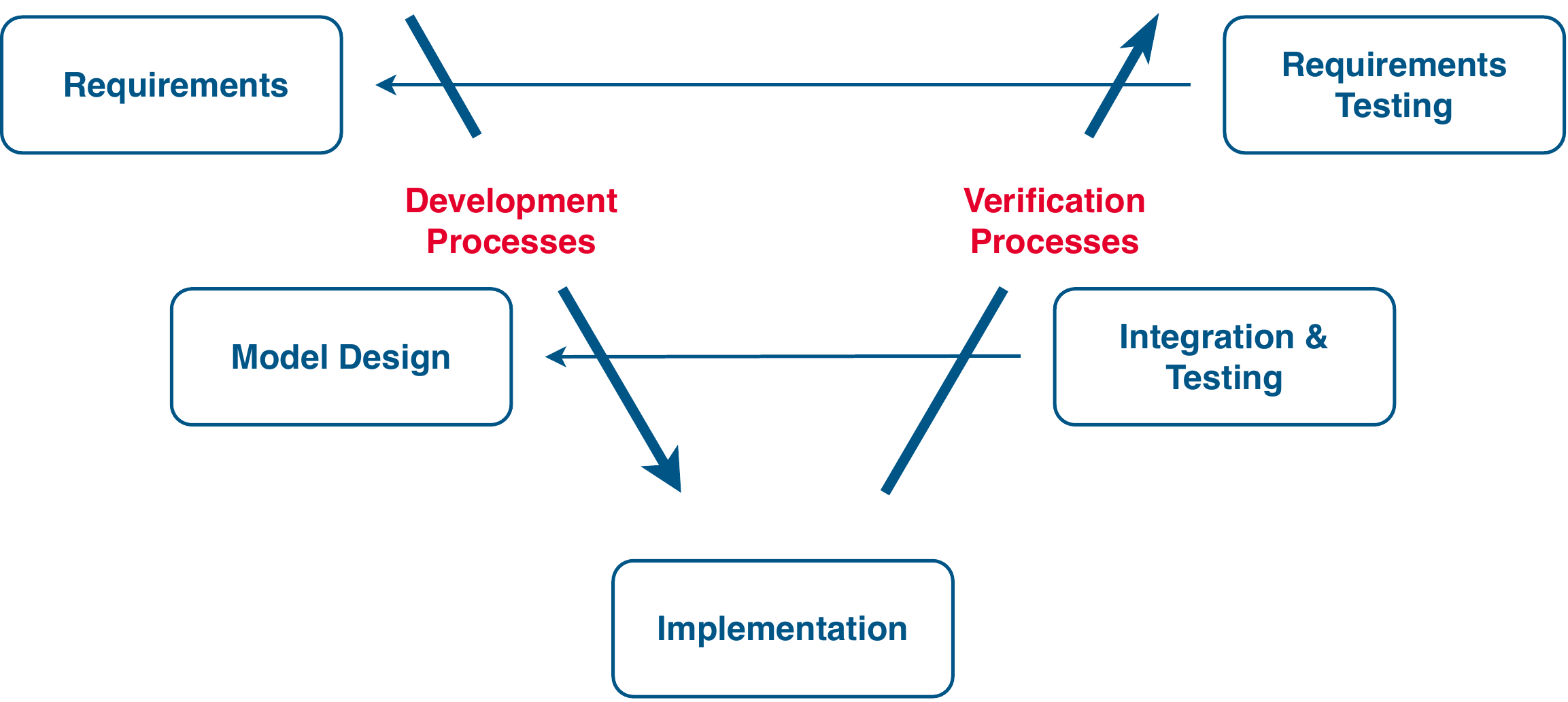}
    \caption{Item development workflow}
    \label{fig:sys_wkflow}
\end{figure}

The item's development workflow is usually represented by the V-cycle (refer to Figure \ref{fig:sys_wkflow}). Today, the usual development paradigms are Requirement-based Engineering (RBE) or Model-Based Engineering (MBE). Indeed, avionic items are developed using \emph{classical programming}, i.e. the developers explicitly use language instructions to implement requirements (or model) and thus are able to control each line of code of the embedded software item.

Recently, new avionic functions have emerged, aiming at developing new flight experiences: navigation/surveillance assistance (e.g. vision-based navigation, obstacle sensing, virtual sensing), speech-to-text applications, autonomous flight, predictive maintenance, cockpit assistance...

Contrary to \emph{classical programming} which can hardly support these functions, Machine Learning (ML) which is a sub-domain of Artificial Intelligence (AI), is well known to show good results for most of them
(e.g. \cite{bahdanau2015, devlin2019, redmon2016}).
Current industrial guidance have a strong focus on bespoke technologies in aeronautical applications thus, are not appropriate to support this paradigm change.

Indeed, ML techniques introduce a brand new paradigm in avionic development: the data-driven design of models (including supervised, unsupervised and reinforcement learning).
\emph{Data-driven} refers to the fact that the data rule the algorithm behavior through a learning phase.
We choose to restrict our research to \textbf{offline supervised learning}, i.e. the training is done with \emph{labelled data} and \emph{before embedding the ML software}.
As a first step to the qualification of a ML item, these restrictions seems reasonable.
Besides, these restrictions do not affect our capability to find efficient algorithms that might solve the emerging functions mentioned above.

Even with this restriction, many significant issues remains regarding the current demonstration of conformity. Indeed, some fundamentals of the usual techniques (Requirement Based Engineering or Model Based Engineering) are jeopardized, challenging the classical safety guarantee argumentation:
\begin{itemize}
  \setlength\itemsep{-0.05em}
  \item Specifiability: It sounds difficult, a priori, to capture the complete behavior of a ML model (for instance when the training dataset is composed of millions of images), decreasing the confidence that the model behaviour will always match the functional intent and be free of unintended behavior that may jeopardize the safety.
  \item Traceability: The relationship between item requirements and corresponding \emph{learned parameters} of the ML algorithm cannot be established. This makes the ML item design less transparent and draw a need for explainability capabilities to add confidence that the algorithm correctly and safely implements the intended function.
  \item Robustness: Perturbations in the design phase and the operational inference of the trained algorithm may lower the reliability on the predictions correctness and may degrade the functional and safety performances.
 \end{itemize}

\begin{figure}
    \centering
    \includegraphics[width=\textwidth]{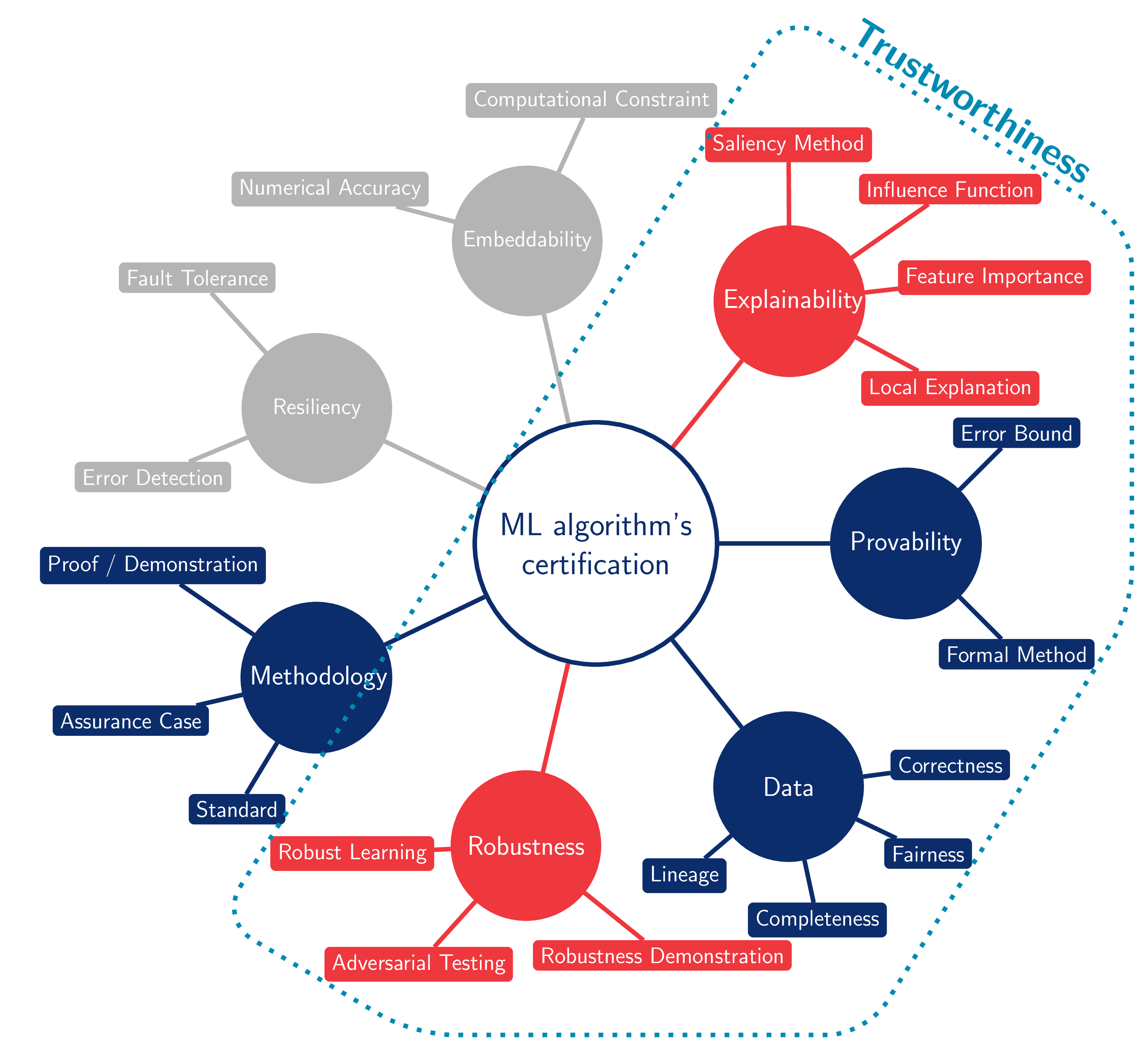}
    \caption{AI certification key elements. In red, are the domain that we detail in this survey. In blue are the domain that we briefly present. In grey are the domain out of our scope.}
    \label{fig:keyelcertif}
\end{figure}

The motivation of this survey is two-fold: {\it (i)} identify the main challenges to the certification of ML based item (see Figure \ref{fig:keyelcertif}), {\it (ii)} overview the literature to know whether existing methods or techniques are able to solve the problems raised by the identified challenges.
Compared with the survey of \citet{huang2018survey}, which is also motivated by the certification of critical embedded systems that include Deep Neural Network (DNN), our survey is guided by aeronautics regulations and standards and is not limited to DNN, although many of the cited works deal with DNN.

Currently, the challenges for certifying ML algorithms are listed in Figure~\ref{fig:keyelcertif}.
The goal of this survey is not to provide a complete overview of possible techniques to solve these challenges, its scope is reduced to the trustworthiness and methodology considerations.
Specifically, the survey focuses on Explanability and Robustness (more precisely adversarial robustness) because they address key issues introduced by the intrinsic characteristics of ML algorithms: the lack of transparency of the design (black-box aspect) and the lack of performances reliability when inputs are slightly perturbed in the range of the Operational Design Domain (ODD).
In grey, in Figure~\ref{fig:keyelcertif}, are the topics that are out of our scope: \textit{Embeddability} and \textit{Resiliency}.
\textit{Embeddability} raises classical implementation concerns such as the deterministic behavior of the item versus memory and computational constraints.
Some resolution techniques are specific to ML, such as the optimization of the ML algorithm structure or implementation architecture to fit the targeted host platform.
\textit{Resiliency} monitors the behavior of the system when it is deployed, e.g. error detection or fault tolerance. This is an issue to be addressed at the system level of engineering.

This paper is structured as follows: in section \ref{sec:certif}, we overview the main challenges raised by the use ML in the demonstration of compliance with regulation requirements.
In section \ref{sec:domains} we introduce the Trustworthiness considerations which will possibly help to fill the existing gaps in future certification approaches (see Figure~\ref{fig:certifOverview}).
Then the sections \ref{sec:robustness} and \ref{sec:XAI} focus on robustness and explainability, respectively.
Eventually, we end the paper with section \ref{sec:directions} which recaps the identified problems for the qualification of ML items.

\section{Impact of ML techniques on certification approach}
\label{sec:certif}
\begin{figure}[b]
    \centering
    \includegraphics[scale=0.11]{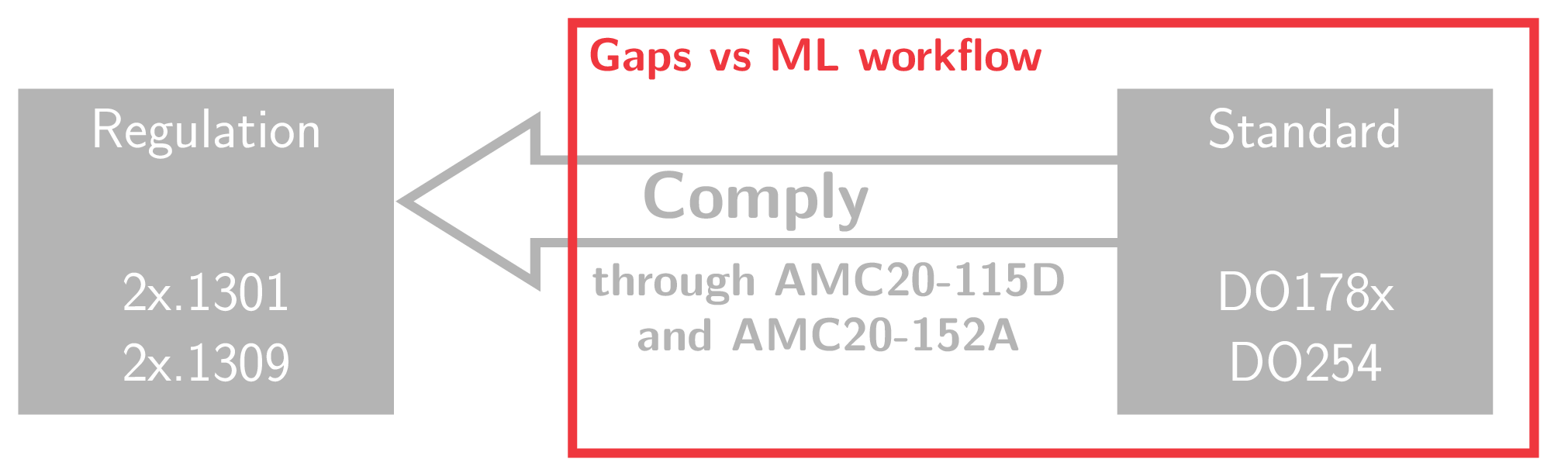}
    \caption{Certification Process Overview}
    \label{fig:certifOverview}
\end{figure}

The European Union Aviation Safety Agency (EASA) provides regulatory material that defines and explains all the requirements due for developing safe avionic products \cite{CS-25}.
Regarding software/hardware items embedded in avionic safety-related systems, the certification specification are described in CS~2x.1301 and CS~2x.1309.
Roughly speaking, we would summarize these paragraphs as follows: \emph{avionic systems should safely perform their intended function under all foreseeable operating and environmental conditions.} Through supplemental documents (AMC20-115D and AMC20-152A\footnote{AMC stands for Acceptable Mean of Compliance}), EASA recognizes the current avionic standards as acceptable means of compliance to the regulation text.
The grey element in Figure \ref{fig:certifOverview} shows the relation between regulations and standards.

Assuming the ML techniques are reduced to offline learning, we do not anticipate any change to the regulation \cite{CS-25}. However, even if regulation requirements are unchanged, the current standards do not provide sufficient guidance to make a complete demonstration of conformity for a ML item. This section details these gaps.

\paragraph{Specifiability} One of the fundamentals of the RBE (or MBE) relies on the correct and complete capture of the item requirements: either they come from system allocated requirements (intent) or from its own behavior (emerging functions). In this context, the item can be verified to safely perform the intended function under all foreseeable operating and environmental conditions.
With ML items, it sounds difficult to fully specify the function with a classical requirement process. For instance, all the possible way to describe a runway whatever the operational (e.g. sensors own bias) and the environment conditions (e.g. weather, light conditions) cannot be defined using textual or even modelling technique. This is the reason why we will prefer use millions of images instead, betting that the data experience will complete the requirement capture and that ML technique will enable the development of an acceptable runway detection function. Therefore the difficulty will be to describe as thoroughly as necessary the operational domain (ODD) in which the detection function is supposed to operate and then to
validate that the collected data correctly and sufficiently represent this ODD.

\paragraph{Traceability} The current software standard requires the traceability between the requirements and the code (in both senses) so that each line of embedded code can be justified as implementing captured requirements.
In a ML development context, this relationship that makes the design a white-box process, is lost.
Actually the trained algorithm is a complex mathematical expression (made of arithmetic operations with weights and bias)  that is not traceable to any upward functional requirements.
Thus it becomes infeasible to demonstrate the completeness of implementation using traceability.
From this lack of traceability comes the loss of transparency of the model, making
the link from the input data to the output predictions not understandable by a human. This lowers the confidence that the safety properties of the intended function are preserved in its operational domain.
In this context, explainable AI could be a means to enhance the confidence in these algorithms when safety is highly critical (see section~\ref{sec:domains} and \ref{sec:XAI}).

\paragraph{Unintended behaviour} All the life-cycle processes (requirements capture, model design, implementation, integration and requirement-based testing) of the item development workflow (see Figure \ref{fig:sys_wkflow}) are used to demonstrate the consistency with the intended function.
In addition to the problems mentioned above, the learning phase could introduce some unexpected behavior that we cannot measure or prevent using usual activities.
It means that we cannot guarantee the absence of unintended behavior during item operation and specifically those affecting aircraft safety.
This may be due to several root causes such as a low-quality data management process (e.g. bias, mislabeled data) or inadequate learning process.
We will see in the rest of the paper that there are methods to limit or to formally verify the occurrence of such effects.

\section{Trustworthiness Considerations}
\label{sec:domains}
The qualification of ML items in the frame of avionic developments offers lots of research challenges in various domains. We selected the overview of 5 different domains that seem very promising and which will contribute to support the certification of ML-based systems.
Thus, this section overviews the domains covering the Trustworthiness (see Figure \ref{fig:keyelcertif}) and the Methodology aspects.

Explainability and robustness of Machine Learning model have already been reviewed, the reader may therefore refer to \cite{huang2018survey, biggio2018, ren2020}, to get more details on these topics.
Nevertheless, the originality of our work stands in the perspective of the conformity to the regulation requirements inherent to the development of safety-critical systems for avionic domain.

\paragraph{Explainability} Explainability is a topic inherent to ML technique.
It is not needed for classical programming since the item development can be fully explained through the joint requirement and traceability processes which enable the interpretation of each line of embedded code. ML development breaks this understanding chain and make room for an undesirable black box effect.
This is the reason why one can think that, when required by the safety level of the implemented function, explainability would be requested to add the necessary confidence to support the demonstration of conformity \cite{roadmapEASA2020}.
\citet{phillips2020} introduce four principles of explainable AI: explanation, meaningfulness, explanation accuracy, knowledge limits.
Basically, these principles state that an explainable AI must: {\it (i)} provide an ``evidence or reason for all outputs'' {\it (ii)} be meaningful with respect to the audience, {\it (iii)} be representative of the way that the algorithm produce the output and {\it (iv)} be aware of the domain of usage of the algorithm, i.e. it should not give an explanation when the input is out of scope since the algorithm is not design to work with it.

In the aeronautical context, four stakeholders could need explanation: the designer, the authorities, the pilot and the investigator.
Considering the ``meaningfulness'' principle, each of them could receive different explanation.
Indeed, the needs for explanation would be different for an engineer who knows the system and for an end-user who has no prior knowledge of the system.
The explanations could also differ whether it is used for debugging purposes during the development (for designer), for investigation purposes in case of in-flight issues (for authorities or investigator) or for a user assessment of the model predictions when the system is deployed, e.g. a pilot who requests justification of the decision to enhance his confidence in the system before acting.
In the literature, we find two kinds of explanation: local explanation \cite{guidotti2018, ribeiro_2016, ribeiro2018, dabkowski2017, fong2017, hendricks2016, kendall2017, zeiler2014} and global explanation \cite{lundberg2017, shrikumar2017, zhao2019} (see Section \ref{sec:XAI}).

\paragraph{Adversarial Robustness} Robustness comprises several sub-domains (adversarial examples, distributional shift, unknown classes, \emph{physical} attacks, ...) but we focus only on \emph{adversarial robustness} which is defined as the capability of algorithms to give the same outputs considering some variation of the inputs in a region of the state space.
The issues addressed by the adversarial robustness domain are at the heart of the certification process; it addresses partially the demonstration of the intended function.
The aim is to assess and/or enhance the behavior of the algorithm when dealing with \emph{abnormal inputs} (noise, corner case, sensor malfunction...).
The literature is two sided: one side works at enhancing the adversarial robustness of the algorithm \cite{carlini2019, ford2019, goodfellow2014, kurakin2016scale, madry2018, papernot2015, papernot2018, szegedy2013, zhang2019}, the other side at its verification \cite{cisse2017, gehr2018, huang2017, katz2017, koh_2017, mirman2018, salman2019, singh_2019, tsuzuku_2018}.
Improving the adversarial robustness consists in finding robust learning procedure that is resilient against crafted exampled made to defeat the ML algorithm.
The verification of this latter property is either based on formal methods  \cite{cisse2017, koh_2017, salman2019, tsuzuku_2018, katz2017} or optimization methods \cite{gehr2018, huang2017, mirman2018, singh_2019}.

\paragraph{Provability} The provability aspect is the capability to formally demonstrate that system properties are preserved. Formal methods provides mathematical evidences to support such demonstration.
As stated above, formal methods are already used to verify well-defined properties of algorithms \cite{katz2017, Katz2019TheMF, wang2018}, such as the robustness \cite{huang2017, gehr2018, mirman2018, singh_2019}.

The error (or generalization) bound gives guarantees on the ability of the algorithm to generalize..
An algorithm generalizes well when it maintains its high performance on unseen data, i.e. the theoretical error, $r$ is close to the empirical error $\hat{r}_{S}$ (computed from a training set $S$).
The idea is to bound the gap between the theoretical error $r$ and its empirical counterpart $\hat{r}_{S}$.
However, $r$ is not computable since it is the error on all possible data.
Hence, the need for probabilistic bounds which give an upper bound on the gap between $r$ and $\hat{r}_{S}$.
The first generalization bounds appears with the PAC (Probably Approximately Correct) theory \cite{valiant1984} and have the following forms:
\begin{align}
    \text{Pr}_{S}\Big(|r - \hat{r}_{S}| \le \epsilon(\text{model's complexity},\# S, \delta )\Big) \ge 1 - \delta,
\end{align}
where $\epsilon(\cdot) \ge 0$ and $\delta \in \left]0,1 \right]$.
$\epsilon$ is a function that models the upper bound that usually relies on the complexity of the model, the number of samples in $S$ (denoted by $\#S$), and the probability $\delta$.
The ideal scenario is to have the gap between $r$ and $\hat{r}_{S}$ small while having a high probability that the inequality holds, i.e. having $\epsilon(\cdot)$ and $\delta$ as small as possible.

Particularly, the PAC-Bayesian theory\cite{mcallester1999,shawe-taylor1997}, which interprets an algorithm as a \emph{majority vote}, is well known to provide tight generalization bound \cite{hernandez2012}.
A majority vote is defined as the weighted sum of several models where the weights of the sum constitute a distribution $Q$.
Hence, a model follows a prior distribution $P$ before the learning and posterior distribution $Q$ after.
A PAC-Bayesian generalization bound could have the following form:
\begin{align}
    \label{eq:error_bound}
    & \text{Pr}_{S}
    \Big(
      r \le \hat{r}_{S} + \epsilon(KL(Q \| P), \# S, \delta)
    \Big)
    \ge 1-\delta.
\end{align}
One can notice that both bounds have a similar structure but the particularity of the PAC-Bayesian theory is to use the Kullback-Leibler divergence between $Q$ and $P$ to quantify the complexity of the model.
Besides, the differences stand also in the definition of $r$ and $\hat{r}_{S}$ since in the PAC-Bayesian theory the algorithm is interpreted as a \emph{majority vote}.

\paragraph{Data management}
In the avionic context, data has been used for a long time with the use of databases or configuration files.
However Machine Learning techniques are bringing a totally new aspect in the qualification approach.
Contrary to \emph{classical programming}, ML design techniques are data-driven, thus the data management process is essential to the demonstration of conformity.
As already stated, building a ML algorithm of good quality, requires data of good quality (no erroneous data, no mislabelled data, ...) but this is not sufficient.
Indeed the data representativeness plays also a significant role.
Representativeness comes from statistics and is a quite challenging problem: it allows to check if the data is a correct snapshot of the phenomenon to be learnt.

The quality of the data can be measured using existing metrics: accuracy, consistency, relevance, timeliness, traceability, and fairness are some of them \cite{cai2015,picard2020}.
\textbf{Accuracy} checks if the data is well measured and stored.
\textbf{Consistency} verifies if the preprocessing of the data does not compromise their integrity.
\textbf{Relevance} measures if there are sufficient data to learn the intended function and if the data contain the correct information regarding the needs.
\textbf{Timeliness} concerns the availability of the data over time.
\textbf{Traceability} verifies the reliability of the data source and if all activities for the transmission do not alter the data integrity.
\textbf{Fairness} is about avoiding undesirable bias in the dataset.

\paragraph{Methodology}
The \emph{methodology} structures the assurance activities that are needed to support the qualification aspects, i.e. the demonstration to the Authorities that the item development is compliant with the standardized guidance.
The red square in Figure \ref{fig:certifOverview} highlights that changes are required in the AMCs and/or the industrial standards.
Therefore it will be necessary to find alternative means to comply with the regulation material, either by adapting the current certification approach or by building a new one.

\begin{figure}
    \centering
    \includegraphics[width=\textwidth]{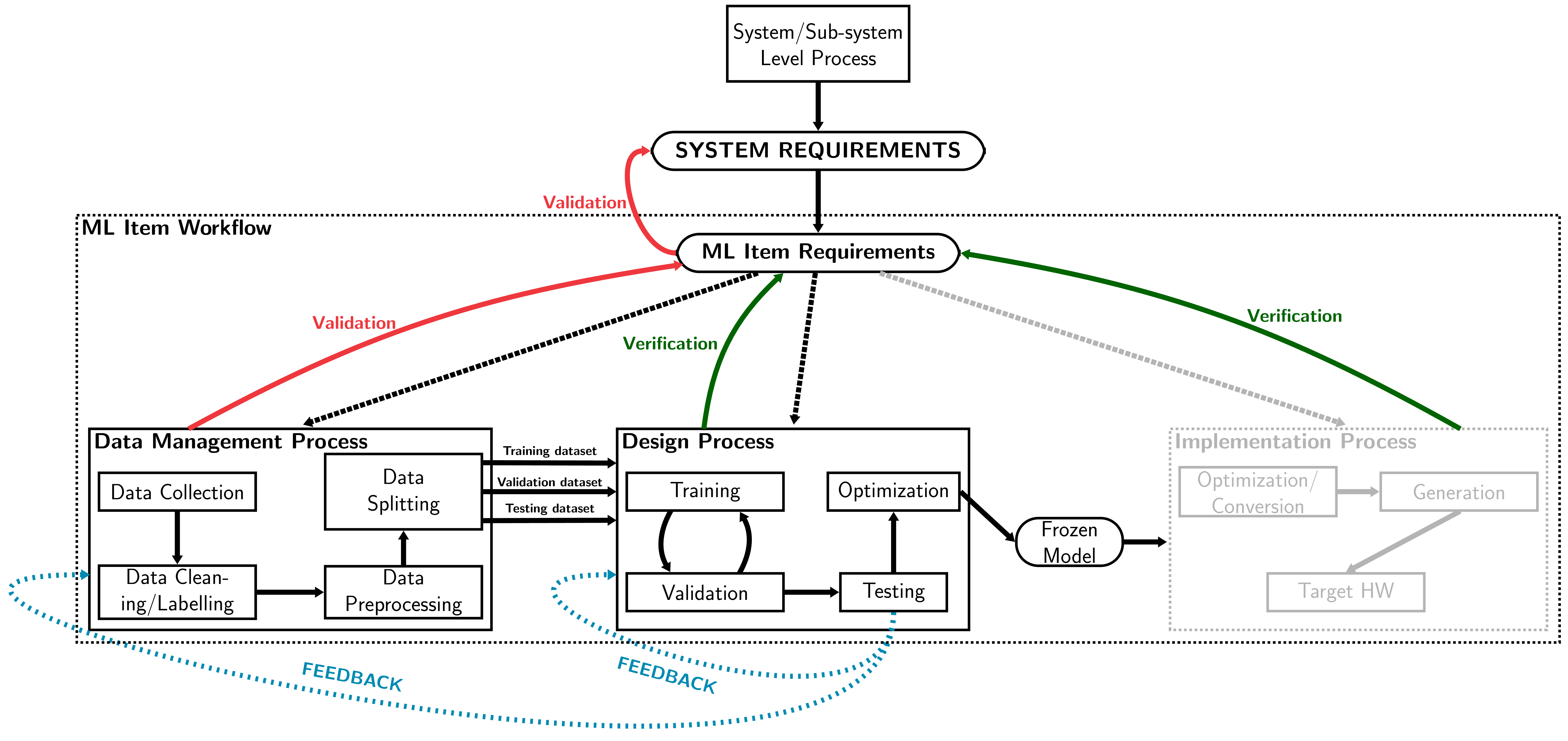}
    \caption{Machine Learning development process: The three main stages of ML development process (Data Management Process, Design Process and Implementation Process). The stage in grey is out of the survey scope. The red arrows show the validation activities. The green arrows show the verification activities. The blue arrows show where we can loop back if the model underperforms.}
    \label{fig:ML_workflow}
\end{figure}

Then it will be crucial to clearly define the development process of ML items (\cite{baylor2017, ashmore2019, amershi2019, studer2020}) to {\it (i)} ease their development and maintainability, and {\it (ii)} identify the validation and verification activities.
Figure \ref{fig:ML_workflow} describes this development process: requirements from the system/subsystem level are refined into ML item requirements to fit to the three main stages of the workflow: Data Management process, Design process and Implementation process.\\
\textbf{Data Management Process} First data are collected with respect to the problem to be solved.
Then, the data should be cleaned and labelled\footnote{as the scope is reduced to offline supervised learning, labelling the data is necessary.}, i.e. inaccurate data samples are removed and each remaining data sample is assigned with a true label (known as ``ground truth'').
Finally, data are preprocessed and split.
Preprocessing encompasses all the tasks that transform the data into a more suitable format for the design phases (e.g. feature engineering or data normalization).
Preprocessing effort may vary depending on the data collection phase (e.g. data comes from different sources) and the type of data (e.g. images, time series, tabular, sound, text). After the preprocessing step, the data is split into a training, a validation and a testing dataset.\\
\textbf{Design Process} The design process takes as input the three datasets and outputs a frozen model.
The processes in this stage are more iterative than sequential.
To tune the hyper-parameters of a model, you iterate between the training and the validation phase.
After the tuning part, you are able to test your model.
If the performances are lower than expected with respect to the ML item requirements, it is possible to loop back either to the data management process or the design process.\\
\textbf{Implementation Process} Within the implementation stage, the specificity of the hardware is taken into consideration: the frozen model is optimized and converted with respect to the host platform constraints and the operational requirements cascaded from the system level. Finally, a binary code is generated and loaded in the hardware target.
Since we only consider the item design aspects, the implementation stage is out of the scope of this survey.

Since its creation in the 80's, the ED-12/DO-178 standard issues gather the \emph{industrial best practices} that impose the necessary rigor of development to avoid the introduction of errors that may lead to a system failure, and thus increase the safety of the system.
Historically, methodologies undeniably brought efficient results in terms of safety.
Thus methodological considerations are key to build a correct certification approach.
One lead to build relevant safety argumentation is the concept of safety case~\cite{rushby_2015} or assurance case \cite{MITRE_2017}.
Indeed, \citet{rushby_2015} argues that the introduction of this kind of methodology (i.e. not prescriptive approaches) in the industries helps to significantly reduce the number of accidents and deaths.
Assurance cases, which are a generalization of safety cases, are defined by MITRE \cite{MITRE_2017} as follows:
\emph{a documented body of evidence that provides a convincing and valid argument that a specified set of critical claims regarding a system's properties are adequately justified for a given application in a given environment}.
Hence, assurance cases could be used during the development process of a ML-based item to build, share and discuss sets of structured arguments to support the demonstration of conformity based on outcomes of specific assurance techniques.

\section{Adversarial Robustness}\label{sec:robustness}

\begin{table}
  \caption{Summary of robustness method: The top of the table shows the papers that provide an attack and\slash or a defense and the bottom shows the papers that provide a verification method.
  The $\ell_p$-norm column refers the norm originally considered in the papers to compute the distance between $x$ and $x_{adv}$\protect\footnotemark.
  ``-'' means that the method do not rely on an $\ell_p$-norm.
  The color code indicate whether the norm is used for attack or defense; black means it is used for both. The colunm ``\# parameters'' reports the size of the Neural Network (in terms of numbers of parameters) used in the papers; it give an insight on the scalability of the associated method.}
  \label{tab:robMethod}
  \begin{tabular}{cccccc}
    \toprule
    \multicolumn{5}{c}{\textbf{Attack/Defense}} \\
    Method Name    &  Papers    &  \textcolor{darkblue}{Attack}  &  \textcolor{darkgreen}{Defense}  & $\ell_p$-norm  \\
    \midrule
    \CW                    &  \citet{carlini2017}      & \checkatt &           &  \attack{\lzero \slash \ltwo \slash \linf}  \\
    \ParsevalTraining       &  \citet{cisse2017}        &           & \checkdef &  \defense{\ltwo}   \\
    \FGSM                  &  \citet{goodfellow2014}   & \checkatt & \checkdef &  \linf   \\
    \IFGSM                 &  \citet{kurakin2016scale} & \checkatt & \checkdef &  \linf   \\
    {\scriptsize\sc gradient-based attack+}{\sc ila}~{\tiny (\cite{huang2019})}  &  \citet{li2020} & \checkatt$^*$ &  &  \attack{\linf}   \\
    \PGD                    &  \citet{madry2018}        & \checkatt & \checkdef &  \linf \slash \attack{\ltwo}   \\
    \deepfool               &  \citet{moosavi2016}      & \checkatt &           &  \attack{\ltwo}  \\
    \JSMA                   &  \citet{papernot2015}     & \checkatt &           &  \attack{\lzero}  \\
    \DKNN                   &  \citet{papernot2018}     & \checkatt$^*$ & \checkdef &  \attack{\linf}\slash\defense{\ltwo}   \\
    \LBFGS                  &  \citet{szegedy2013}      & \checkatt & \checkdef &  \ltwo   \\
    \LMT                    &  \citet{tsuzuku_2018}     &           & \checkdef &  \defense{\ltwo}   \\
    \GDA                    &  \citet{Zantedeschi2017}  &           & \checkdef &  -       \\
    \YOPO                   &  \citet{zhang2019}        &           & \checkdef & \defense{\linf}    \\
    \midrule
    \multicolumn{5}{c}{\textbf{Verification}} \\
    Method Name             &  Papers                     &  Sound     &  Complete  &  \# paramaters \\
    \midrule
    \aiai                   &  \citet{gehr2018}           & \checkmark &            & $>$ 238,000 \\
    \DLV                    &  \citet{huang2017}          & \checkmark & \checkmark &   $>$ 138,000,000\\
    \reluplex               &  \citet{katz2017}           & \checkmark & \checkmark & $\sim$ 13,000   \\
    \marabou                &  \citet{Katz2019TheMF}      & \checkmark & \checkmark & $\sim$ 13,000   \\
    \refinezono &  \citet{singh_2019}         & \checkmark &            &  $>$ 1,000,000 \\
    \reluval                &  \citet{wang2018}           & \checkmark &            &              \\
  \bottomrule
\end{tabular}

\vspace{0.2cm}
\small{$^*$ The authors adapt existing attacks to target the weakness of their defense.\hfill}
\end{table}
\footnotetext{It is important to note that the techniques can be adapted with different $\ell_p$-norm}

Concerning the fundamentals of the regulation requirements, i.e. mainly to demonstrate that the system safely performs the intended function, the key role of robustness is twofold:
on one hand and according to ED-12C/DO-178C, it is \emph{the extent to which software can continue to operate correctly despite abnormal inputs and conditions.}
On the other hand and more specifically to ML application, \citet{roadmapEASA2020} states that ML system is robust when it \emph{produces the same outputs for an input varying in a region of the state space}.
Perturbations can be natural (e.g. sensor noise, bias...), variations due to failures (e.g. invalid data from degraded sensors) or maliciously inserted (e.g. pixels modified in images) to fool the model predictions.
When perturbed examples fool the ML algorithm we talk about \emph{adversarial examples}.
It is commonly defined as noises on inputs that are imperceptible or that do not exceed a threshold.
We state in Definition~\ref{def:adversarial-example} two possible manners of generating adversarial examples, \ie attacking an ML model.
\begin{definition}{\textit{Adversarial Example.}}
\label{def:adversarial-example}
Let $(x,y)$ be a benign example and the true label, $\delta$ be a perturbation, $\epsilon$ be the maximum allowed perturbation, $L$ be a loss function and $f$ be the trained ML model, we have\newline
\begin{minipage}{0.45\textwidth}
\begin{align}
    \label{eq:adv-ex-min-pert}
                           &\min \normp{p}{\delta}     \nonumber \\
    \text{such that} \quad &f(x+\delta) \neq y, \\
                           &x+\delta \in Domain(x) \nonumber
\end{align}
\end{minipage}
\hfill or \hfill
\begin{minipage}{0.45\textwidth}
\begin{align}
    \label{eq:adv-ex-max-loss}
                           &\max L(f(x+\delta), y)     \nonumber \\
    \text{such that} \quad &\normp{p}{\delta} \leq \epsilon, \\
                           &x+\delta \in \text{Domain}(x) \nonumber
\end{align}
\end{minipage}
~\\
where $\normp{p}{\cdot}$ is an $\ell_p$-norm and Domain($x$) depict the allowed values for the example $x$.
\end{definition}

As stated in Definition~\ref{def:adversarial-example} the optimization problem leads to get an adversarial example with a different label than the original one; we refer to it as ``untargeted attack''.
Few modifications of the optimization problem allow to have a ``targeted attack'', i.e. the possibility to choose the label of the adversarial examples.
Let $t$ be the targeted label we have,\\
\begin{minipage}{0.5\textwidth}
\begin{align*}
                           &\min_\delta \normp{p}{\delta}\\
    \text{such that} \quad &f(x+\delta) = t, \\
                           &x+\delta \in Domain(x)
\end{align*}
\end{minipage}
or \hfill
\begin{minipage}{0.5\textwidth}
\begin{align*}
                           &\min_\delta L(f(x+\delta), t)\\
    \text{such that} \quad &\normp{p}{\delta} \leq \epsilon, \\
                           &x+\delta \in \text{Domain}(x)
\end{align*}
\end{minipage}.

Moreover, the attacks could be done in either a white-box or black-box setting.
White-box setting basically means that the \emph{attacker} has full access to the model, its parameters and the training dataset while black-box setting means that the \emph{attacker} can only query the model with data.

Finally, the means deployed to overcome adversarial attacks is known as adversarial defenses.
One of the most efficient defense is the Adversarial Training (AT) \cite{goodfellow2014, kurakin2016scale, madry2018}.
It consists in augmenting your training dataset with adversarial examples.
However, it is often observed that the adversarial training based on a particular $\ell_p$-norm are less effective against attacks based on different $\ell_p$-norm.

Nevertheless, the defenses only enhance the adversarial robustness of ML algorithms whereas for the demonstration of conformity, we will need proof that systems based on the ML algorithm behave safely and as expected.
Actually, there exists methods in the literature that verify that an ML algorithm is adversarially robust.
It could bring the guarantee needed to demonstrate the conformity to the regulation requirement.
Thus, to handle the adversarial robustness issue, we review adversarial attacks/defenses (cf.~Section~\ref{subsec:adv_tech}) and verification methods (cf.~Section~\ref{subsec:verif}).
We report in table~\ref{tab:robMethod} adversarial attacks/defenses and verification methods.
It is important to highlight that, for the attack/defense methods, we report the $\ell_p$-norm used in the paper.
However the methods could be adapted to other norms.

\subsection{Adversarial attacks/defenses}\label{subsec:adv_tech}
\citet{szegedy2013} was one of the first to point out adversarial examples as a weakness of ML algorithm.
The highlighting of this weakness gave rise to numerous researches around adversarial attacks and defenses \cite{ford2019, kurakin2016scale, gourdeau2019, szegedy2013, papernot2015, goodfellow2014, koh_2017, kurakin2016physical, madry2018, papernot2018, zhang2019}.
\citet{carlini2019} provide advice and good practices about the evaluation of the adversarial robustness.
Especially, they claim that one that develops a new defense mechanism must think about an ``adaptive attack'' to evaluate the efficiency of their defense mechanism.
In other words, to evaluate the defense efficiency of your new defense mechanism, you have to test the worst attack against it.
\subsubsection{Attack}\label{subsubsec:attack}

\citet{goodfellow2014} develop an efficient method to find adversarial example called Fast Gradient Sign Method (\FGSM).
The attack consists in crafting the adversarial example \xadv, by adding a fraction, $\epsilon$, of the loss gradient's sign with respect to the input to the original example $x$:
\begin{align*}
    \xadv = x + \epsilon \cdot \text{sign}(\nabla_x L(f(x), y)).
\end{align*}
Besides, the authors show that adversarial examples are invariant to the learning and the architecture, i.e. different architecture trained on different subsets of the dataset misclassify the same adversarial example.
Later, \citet{kurakin2016scale} propose an iterative version of \FGSM~which we denote \IFGSM~where at each iteration, a perturbation is added to $x$ by applying \FGSM~to finally obtain \xadv~after the desired number of iterations:
\begin{align*}
    \xadv^0 = x  \qquad \xadv^i = \xadv^{i-1} + \epsilon \cdot \text{sign}(\nabla_x L(f(\xadv^{i-1}), y)).
\end{align*}

Since perturbations are added several times to $x$ in \IFGSM, the $\epsilon$ chosen is therefore smaller than in \FGSM.
The attack introduced by \citet{madry2018} based on Projected Gradient Descent (\PGD) is similar to \IFGSM~except that \PGD~randomly initialize $x$ before the optimization:
\begin{align*}
    \xadv^0 = x + noise  \qquad \xadv^i = \xadv^{i-1} + \epsilon \cdot \text{sign}(\nabla_x L(f(\xadv^{i-1}), y)).
\end{align*}
\citet{carlini2017} also develop a gradient based method but with a slightly different formulation of the optimization problem (cf. Definition~\ref{def:adversarial-example}).
It is similar to Equation~\ref{eq:adv-ex-min-pert} but at the same time they minimize the margin of the model.
Besides, the authors provide the formulation of their method for three distance metrics (\lzero, \ltwo~and \linf).
\citet{moosavi2016} tailor their attack framework, called \deepfool, for finding the minimum perturbation necessary with respect to the \ltwo-norm\footnote{Note that the authors explain how to adapt their method to other $\ell_p$-norm} to fool the algorithm.
They first simplify the optimization problem to a linear classifier, derive the optimal solution for it and finally adapt the optimal solution found for the linear classifier to neural network.
They provide the algorithm of \deepfool~which describe their iterative approach to estimate the smallest perturbation to create an adversarial example (cf. Algorithm 2 of \cite{moosavi2016}).
Their results show that the adversarial example crafted by \deepfool~is often closer to the original example than the ones crafted by \FGSM~and \LBFGS~\cite{szegedy2013}.

\citet{papernot2015} leverage the ``forward derivative'' of a network $f$, to design the  Jacobian-based Saliency Map Attack (\JSMA).
To obtain the forward derivative, they compute, from the input layer to the output layer, the derivative of the network $f$, instead of the derivative of its loss function, with respect to the input $x$.
It basically corresponds to the Jacobian of the function learned by the network $f$.
Then, the authors derive a ``adversarial saliency map'' based on the Jacobian of the network which points out the input feature that should be modified in order to get a significant impact on the network's output.
We refer the reader to their paper \cite{papernot2015} to get more detail on the algorithm they provide to craft an adversarial example using forward derivative and adversarial saliency map.

\citet{kurakin2016physical} assess the adversarial robustness of models deployed in ``real-world conditions'', \ie the only way to communicate with the systems is through its sensors.
Indeed this is a legitimate question since an attacker will not necessarily have access to the ML model.
In their experiments, the authors feed an image classification algorithm through a camera.
However, they still use the model to generate adversarial examples.
They demonstrate that models are still vulnerable against adversarial attack even in ``real-world conditions''.

In the avionic context, security mesure will be taken to restrict the access to the models and the datasets to undermine the white-box and black-box attacks.
Nevertheless the threat still exists since \citet{li2020} propose an attack in a more restricted setting than black-box known as ``no-box'' setting \cite{chen2017} where the attacker only have very few number of examples that are not training examples.
They succeed to efficiently fool the models trained on the imagenet dataset by developing an auto-encoder called ``prototypical reconstruction'' that manages to learn well with very few data.
An auto-encoder consists of two parts: an encoder and a decoder.
The encoder, encodes the input by learning a new representation of the original input and the decoder strives to reconstruct the input from the encoding as close as possible to the original input.
The authors introduce a new loss for their auto-encoder that is well suited for gradient based attack.
Then, their attack consists in using ILA \cite{huang2019}, which improves the transferability of the attack, in addition to the gradient-based method.

\subsubsection{Defense}
As stated earlier, one of the most effective defense, called Adversarial Training (AT), consists in replacing your original training dataset by an ``adversarial dataset'' crafted with an attack.
However, this method is time consuming, since it requires to generate adversarial example at each step of the learning phase.
Nevertheless, \citet{goodfellow2014} overcome this issue with \FGSM~and propose an adversarial training where they replace a part of the training dataset with perturbed example.
Following this principle, other works with stronger attacks propose adversarial training based on these attacks~\cite{kurakin2016scale, madry2018}.
Besides, \citet{madry2018} show that the adversarial training principle boils down to solve the following min-max optimization (or saddle-point problem):
\begin{equation}
    \min_\theta \EE_{(x,y)\sim \D} \left[ \max_{\normp{p}{\delta} \le \epsilon} L(f_\theta(x+\delta), y)\right],
    \label{eq:minmax}
\end{equation}
where \D~is the unknown distribution of the dataset, $f_\theta$ is the model whose parameters is $\theta$, $L$ is the loss and $\epsilon$ is maximum allowed perturbations.
Indeed, the equation \ref{eq:minmax} comprises two parts: maximization and minimization.
On the one hand, the equation maximizes the loss regarding the noises $\delta \le \epsilon$, \ie it makes $x+\delta$ more likely to be an adversarial example.
On the other hand, the equation minimizes the expectation of the loss regarding the model's parameters.
To the best of our knowledge, AT based on \PGD~is one of the most efficient defense against attacks using the $\ell_\infty$-norm.

Though the methods presented above make AT feasible, it still increase the learning time of a model.
\citet{zhang2019} aim at improving the computation cost of AT by reformulating it as a differential game and then derive the Pontryagin's Maximum Principle (PMP)\footnote{PMP is used in optimal control theory to find the best solution regarding input and constraint.}.
The PMP reveals that AT are closely linked with the first layer of neural networks.
Therefore they develop \YOPO~(You Only Propagate Once) which leverage this fact by limiting the number of forward and backward propagation without worsening the network's performance.
Their experiments show that \YOPO~learn $4$ to $5$ time faster to achieve as good results as adversarial training based on \PGD.

\citet{papernot2018} find another mean to enhance the robustness of ML algorithms.
The authors develop a method called Deep k-Nearest Neighbors (\DKNN), i.e. a hybrid classifier that mixes Deep Neural Network (DNN) and kNN.
Their motivation is to improve the confidence estimation, the model interpretability, and the robustness.
The principle of \DKNN~is to find the nearest neighbors (from the training set) of an input $x$ at each layer of the DNN and find the classes of each neighbor.
This procedure allows the analysis of the evolution of classes in the neighborhood of $x$ throughout the network.
That is why, the authors introduce several metrics such as \emph{nonconformity} and \emph{credibility}.
The nonconformity metric is used to measure the discrepancy between the labels of the neighborhood and the predicted label of an input $x$.
A high nonconformity value means that the labels predicted for the neighborhood of $x$ is different from the one predicted for $x$.
The computation of the credibility measure is based on ``calibration dataset'' whose examples were not used for the training.
Then, the credibility of an input $x$ is the ratio of nonconformity measures of the examples from the calibration dataset that are greater than the input's nonconformity measure.
\DKNN~increases the \emph{confidence} in predictions thanks to the \emph{credibility measure} which is used to assess and select model's predictions.
Moreover, this algorithm becomes more interpretable because of the neighborhood it provides which gives an insight of its ``internal work''.
\citet{papernot2018} claim that \DKNN~would prevent adversarial examples by assigning to them a low credibiility measure.
Their empirical results show particular encouraging results against \CW~attack.

The drawback of the above techniques lies on their dependence to the $\ell_p$-norm.
Indeed, even if it brings robustness against attacks relying on the same metric distance (like \PGD), the defense could become ineffective when the attacks depend on a different metric distance.
Instead of searching for the perturbation that maximizes the loss function of a model (Equation \ref{eq:minmax}), \citet{Zantedeschi2017} propose to consider all the local perturbations (drawn from a gaussian distribution) around each examples:
\begin{align*}
    \min_\theta \EE_{(x,y)\sim \D} \; \EE_{\delta\sim \normal(0,\sigma^2)} L(f_\theta(x+\delta), y),
\end{align*}
where \D~is the unknown distribution of the dataset, $f_\theta$ is the model whose parameters is $\theta$ and $L$ is the loss function of the model.
They claim that their method, known as Gaussian Data Augmentation (\GDA), instead of crafting adversarial example only in the direction of the gradient, explore much more directions around the examples.
\GDA~outperforms AT considering adversarial accuracy measure (\ie accuracy on a attacked dataset) on MNIST~\cite{lecun1998} and CIFAR-10~\cite{krizhevsky2009}.

A hypothesis that could explain the adversarial phenomenon is the high expressiveness of the models, \ie the the capacity of the model to learn complex behavior.
\citet{cisse2017} and \citet{tsuzuku_2018} explore this lead by constraining the Lipschitz constant of neural networks to enhance their robustness.
Considering all possible couple of points of the input space of a function $f$, the Lipschitz constant is the smallest value that upperbound the absolute value of the slopes given by each pair of points.
Intuitively, we would say that constraining the Lipschitz constant of a function $f$ to a small value would smooth the function and therefore reduce its expressiveness.
\citet{cisse2017} develop specific training called \ParsevalTraining~which consists in ensuring that the Lipschitz constant of all the network's layers is smaller than $1$ by having Parseval tight weight's matrices.
We refer interested readers to the paper \cite{cisse2017} for the mathematical details.
\citet{tsuzuku_2018} propose another training method, called Lipschitz Margin Training (\LMT), to constrain the Lipschitz constant of a network.
They first link the margin of network to the Lipschitz constant.
From that, they derive an algorithm relying on the relation between the margin and the Lipschitz constant that enhance the robustness of the network.
Besides, \LMT~provide a ``certified'' lower bound on the smallest perturbation that can defeat the network.
This latter property could be desirable for the demonstration of conformity to the regulation requirement of an embedded system based on ML.
Indeed, theoretical results are great assets for the certification process.
For example, \citet{gourdeau2019} leverage the \textbf{PAC theory} (Probably Approximately Correct) to provide theoretical proof of the feasibility of robust learning from the perspective of computational learning theory.
The authors focus on the setting where the input space is the boolean hypercube $\mathcal{X} = \{0,1\}^n$ and show that robust learning is feasible or not for different classes of models.

\paragraph{Remark} \citet{ford2019} conduct experiments showing that improving adversarial robustness also improve corruption robustness, i.e. the robustness of the model against distributional shift.
They want to show that both type of robustness could be the manifestation of the same phenomenon.
Nevertheless, it is encouraging that improving adversarial robustness has a positive impact on corruption robustness.

\subsection{Verification}\label{subsec:verif}
The existing frameworks for verifying ML models show that is possible to verify many types of property as long as you are able to express it in the manner expected by the verification framework.
However, the literature focuses on a specific property which is the robustness.
For that reason we decided to distinguish \emph{\hyperref[subsubsec:prop_verif]{\ref{subsubsec:prop_verif} Properties Verification}} and \emph{\hyperref[subsubsec:rob_verif]{\ref{subsubsec:rob_verif} Robustness Verification}}.
Basically, there are two important criteria for a verifier: \emph{completeness} and \emph{soundness}.
\emph{Completeness} means that the verifier can verify all the properties that hold whereas \emph{soundness} means that the verifier cannot prove any wrong property.

\subsubsection{Properties Verification}\label{subsubsec:prop_verif}
Some methods in the literature claim to do \emph{property verification} \cite{katz2017, Katz2019TheMF, wang2018}.
As stated earlier, verifying that the desired properties hold for an ML algorithm presupposes that we can express them as expected by the verifier, \ie in the format that verifier could verify.

\citet{katz2017} develop a method that uses Simplex algorithm ---a method to solve optimization problem of linear programming--- and Satisfiability Modulo Theories solver (SMT solver) ---a solver for formulas of first-order logic  with respect to some background theories such as arithmetic, arrays,~...--- which the authors adapt to work with Neural Network using specific activation function (ReLU: $y = max(0, x)$).
Their method is called \reluplex~and they run it on a case study: ACAS XU system.
They use quantifier-free formulas to the formalization of properties which consists in constrained domains for inputs and outputs.
Basically, they reduce the verification problem into a constraint satisfiability problem (CSP).
They verified 10 properties with \reluplex~and got a timeout for 2 of them, i.e. \reluplex~was not able to end the verification within the given time.
The longest verification took almost 5 days while the fastest took less than 8 minutes.
\marabou~\cite{Katz2019TheMF}, which is the successor of \reluplex, is based on the same principles.
The framework, now supports piecewise linear layer and activation and has a ``divide and conquer mode'' which improves the computation time of the verification.

\subsubsection{Adversarial Robustness Verification}\label{subsubsec:rob_verif}
Robustness verification is an emerging field \cite{cisse2017, gehr2018, gourdeau2019, huang2017, singh_2019, tsuzuku_2018}.
Verifying robustness means ensuring that the algorithm outputs the same label $y$ for an example $x$ and its neighborhood whose the computation usually rely on metric distance (\eg $\ell_p$-norm)
This property could be stated as follow:
\begin{align}
    \label{eq:robustness-property}
    \forall x'\; \text{s.t.}\; \normp{p}{x - x'}\le \epsilon,\quad f(x) = f(x'),
\end{align}
where $\normp{p}{x - x'}\le \epsilon$ represent the neighborhood around $x$ such that the distance between $x$ and any neighbor $x'$ does not exceed $\epsilon$ and $f$ is the ML model.
Indeed, the existence of adversarial examples could be seen as the negation of Equation~\ref{eq:robustness-property}.
Thus, verifying that Equation~\ref{eq:robustness-property} holds boils down to check the adversarial robustness of an algorithm.

\begin{figure}
    \centering
    \includegraphics[scale=0.15]{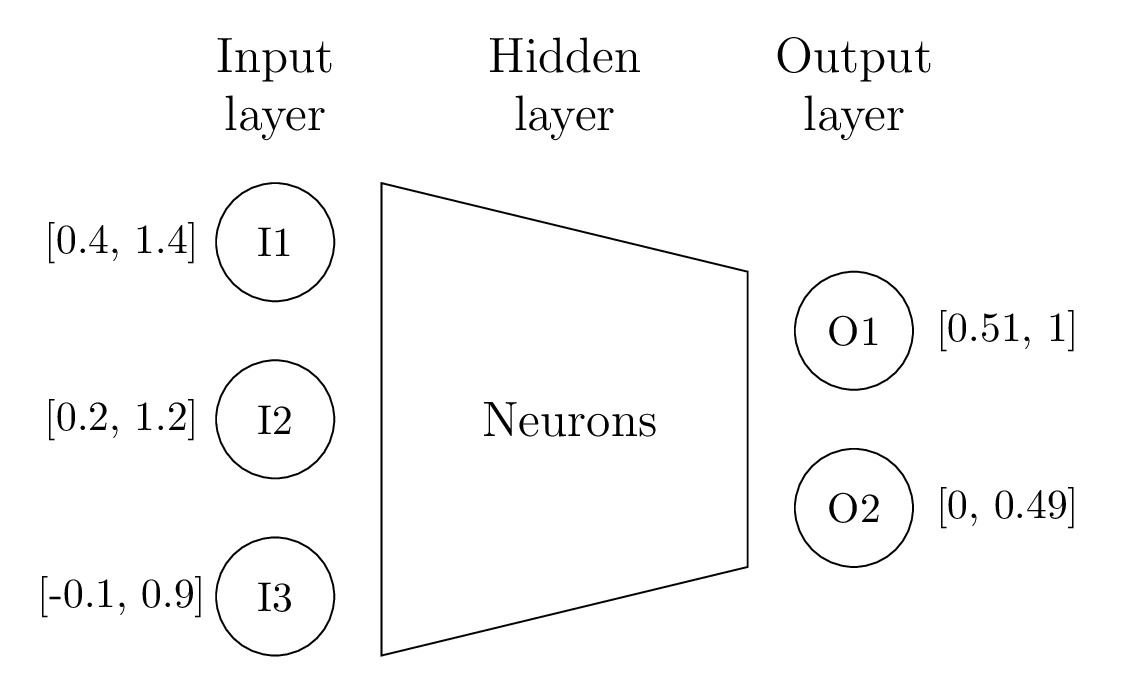}
    \caption{Robustness property example - We consider a hypercube (of size 0.5) as a neighborhood around $x=(0.9, 0.7, 0.4)$ and the output must always be the first class, \ie the lower bound of the output neuron O1 must be greater than the upper bound of the neuron O2.}
    \label{fig:property_example}
\end{figure}

\citet{singh_2019} develop \refinezono, a verification framework mixing optimization techniques (MILP, LP/MILP relaxation\footnote{Mixed Integer Linear Programming / Linear Programming}) and abstract interpretation ---a method mainly use in static analysis that leverage overappromixation to analyse the behavior of computer programs---.
Mixing those techniques allows better scalability for complete verifiers and improves the precision of incomplete ones.
The principle of \refinezono~is to compute the boundaries of the neurons using optimization techniques and abstract interpretation.
As in \cite{katz2017}, a property is expressed as domains for inputs and outputs.
A robustness property expressed as \emph{same label predicted whatever the point within a region around $x$} can be easily transformed into domains for input and output: the region around $x$ is the domain for the input and the domain for the output is given to have the right class.
For example, Figure  \ref{fig:property_example} shows a robustness property where we verify that in the neighborhood of  the point $x = (0.9, 0.7, 0.4)$ the model outputs only the desired class (the output layer predict always the same class).

Instead of trying to scale up to complete verifiers, such as MILP, with abstract interpretation \cite{singh_2019}, it is possible to make abstract interpretation as precise as possible \cite{gehr2018}.
The benefit of applying only abstract interpretation is to verify properties on large neural networks (e.g. CNN).
\citet{gehr2018} develop \aiai~framework which implements abstract interpretation techniques to verify neural networks.
Their framework also focus on the verification of the robustness property (Equation~\ref{eq:robustness-property}).
In \aiai \cite{gehr2018}, the input is replaced by an abstract domain then it is propagated through the network thanks to abstract transformers until the output layer.
Property verification is applied to the abstract domain obtained at the output layer.
The method is sound but incomplete, meaning that for a property the result can either be true or inconclusive.
Hence the choice of the abstract domain for the input is important because it influences the precision of the verifier.

\citet{huang2017} reduce the proof of robustness to a search of adversarial examples.
They focus on the robustness of image classification which is a quite hard problem considering the possible input domain of an image and all perturbation it can have.
The first assumption is that \emph{images are discrete}.
Therefore, \citet{huang2017} explore completely the region of an image $x$, searching for an adversarial example, by defining a set of \emph{minimum} manipulation.
To develop a verification procedure that ensures the safety of a point, the authors first formally define a \emph{safe point}.
Based on their assumption of ``discrete images'', the authors develop an algorithm of \emph{safety verification} using SMT solvers; their tool is called Deep Learning Verification (\DLV).
Since the method of \cite{huang2017} is sound and complete, if the algorithm does not find any adversarial example, it means that there are none for the region and the manipulation set considered.
Moreover, \citet{huang2017} show that their algorithm is scalable to big neural networks by doing experiments on neural networks with more than one hundred million of parameters.

\citet{xu2020} provide a framework, named \lirpa, that unifies the methods of linear relaxation whose goal is to find affine functions that upperbound and lowerbound the output of the network.
For example, to prove a robustness property (as stated in Equation~\ref{eq:robustness-property}) with \lirpa, we must show that the lower bound of the true label is greater than alll upper bounds of all other classes; this would mean that in the worst case, the model will always output the correct class.

\section{Explainability}\label{sec:XAI}

\begin{table}
  \caption{Summary of explanation method. The column ``method name'' report the technique used when the proposed method has no name}
  \label{tab:XAIMethod}
  \begin{tabular}{lccc}
    \toprule
                                        & Method name         & Papers                & Model targeted \\
    \midrule
    \multirow{9}{*}{Local Explanation}  & \maskingmodel      & \citet{dabkowski2017} & \agnostic           \\
                                        & \MPSM              & \citet{fong2017}      & \agnostic      \\
                                        & \LORE              & \citet{guidotti2018}  & \agnostic      \\
                                        & \GVE               & \citet{hendricks2016} & \NN            \\
%                                        & image explanation  & \citet{kendall2017}   & \NN            \\
                                        & {\sc influence function} & \citet{koh_2017}      & \NN      \\
                                        & \LIME              & \citet{ribeiro_2016}  & \agnostic      \\
                                        & \anchors           & \citet{ribeiro2018}   & \agnostic      \\
                                        & \deconvnet         & \citet{zeiler2014}    & \NN            \\

    \midrule
    \multirow{3}{*}{Global Explanation} & \SHAP              & \citet{lundberg2017}  & \agnostic      \\
                                        & \deeplift          & \citet{shrikumar2017} & \NN            \\
                                        & \NPSEM             & \citet{zhao2019}      & \agnostic      \\
  \bottomrule
\end{tabular}

\vspace{0.2cm}
\small{where \agnostic~= model agnostic and \NN~= DNN specific.\hfill}
\end{table}

As already stated earlier, explainability is a very new constraint for embedding ML-based systems.
Implicitly contained in the traditionally programmed components, it has become a true challenge to demonstrate that the ML system's outcomes are trustworthy.
Improving this level of confidence seems inescapable to meet the acceptance criteria of the ML application user (designer, authorities, investigators, pilot...).
It has been clearly identified as a means of acceptance in the EASA AI roadmap \cite{roadmapEASA2020}.

We identify two type of explanations : local explanations and global explanation.
Both subdomains suit different needs: {\it (i)} explanation at the prediction level (local) and  {\it (ii)} explanation of the behavior of the model regarding inputs evolution (global).
In each subdomain, we face two types of explanation techniques: \emph{model agnostic} and \emph{model specific}.
Model agnostic means that whatever the ML algorithm, we can provide an explanation \cite{dabkowski2017, fong2017, guidotti2018, lundberg2017, ribeiro_2016, ribeiro2018, zhao2019}. Model specific means that it will only work with a specific type of ML algorithm; most papers in the literature focus on DNN \cite{ hendricks2016, kendall2017, koh_2017, shrikumar2017, zeiler2014}.

\subsection{Local Explanation}\label{subsec:local_explanation}
Computer Vision is a complex domain because of the nature of the data we have to deal with: images or videos.
Since the development of DNN, it has become easier to reach good performance on image classification, object detection, ...
However, we still do not fully understand why methods based on DNN work so well.
Hence researchers develop new methods to explain, assess, and increase the confidence we can have in the decision taken by DNN for computer vision tasks.

\subsubsection{Model agnostic}
We will first speak of 3 systems: \LIME~(Local Interpretable Model-agnostic Explanations)~\cite{ribeiro_2016}, \anchors~\cite{ribeiro2018} and \LORE~(LOcal Ruled-based Explanation)~\cite{guidotti2018}. Note that \LORE~is not applicable to images. These techniques use inputs and outputs and try to approximate locally the behavior of the ML algorithm.
They are based on the same principle: generate a neighborhood around a sample $x$ and provide an explanation for the sample $x$ thanks to the neighborhood.
If the reader wants to go further, \citet{garreau2020} provide theoretical explanations of \LIME.

The generation of the neighborhood differs according to the system.
\LIME~and \anchors~use an interpretable representation and apply some perturbation on it.
An interpretable representation could be a binary vector where the ones are the relevant information and a perturbation could be the modification on the vector (0 switches to 1) but this should be defined according to the problem that needs to be solved.
\citet{ribeiro_2016} define metrics to measure the distance between the neighbors and the original sample.
\anchors~and \LIME~are using the same method of neighborhood generation.
\LORE~uses a genetic algorithm to generate a balanced neighborhood.
The genetic algorithm is tuned in order to create the best possible neighbors around the desired sample.

Since the neighborhood is generated, it remains to provide the explanation.
In \LIME~linear models are used while in \anchors~decision trees are used.
% Finding anchors involves an exploration or multi-armed bandit problem, which originates in the discipline of reinforcement learning A MODIFIER !!!
For the \anchors~explanation, ``anchors'' are used, i.e. the minimum number of features that lead to the right prediction.
Note that, \citet{ribeiro2018} define several bandit algorithms in order to find the best anchors.
An explanation given by \LIME~correspond to the area of the image that help to take the decision.
\anchors~and \LORE~provide explanation in the IF-THEN form.
\citet{guidotti2018} (\LORE) extract explanations from the decision tree while \citet{ribeiro2018} (\anchors) only check the presence or absence of anchors.
Figure~\ref{fig:lore_example} shows an example of an explanation given by \LORE: the $r$ corresponds to the explanation of the decision and $\Phi$ corresponds to the minimum changes that should occur to flip the decision.
The original decision (0) comes from an algorithm which was trained on the German dataset\footnote{\href{https://archive.ics.uci.edu/ml/datasets/statlog+(german+credit+data)}{https://archive.ics.uci.edu/ml/datasets/statlog+(german+credit+data)}} to recognize good ("0") or bad ("1") creditor according to a set of attributes (age, sex, job, credit amount, duration, ...).

\begin{figure}[t]
\centering
{\bf - LORE} \hfill
\vspace{1mm}
\begin{minipage}[c]{\linewidth}\em
\begin{tabular}{lp{60ex}}
r = & (\{credit\_amount > 836,
      housing = own, other\_debtors =\\ & \quad none,
      credit\_history = critical account\}
       $\rightarrow$ decision = 0)\\
$\Phi$ = & \{ (\{credit\_amount $\leq$ 836,
      housing = own, other\_debtors =\\ & \quad none,
      credit\_history = critical account\}
     $\rightarrow$ decision = 1),\\
  &   (\{credit\_amount > 836,
       housing = own, other\_debtors =\\ & \quad none, credit\_history = all paid back\}
      $\rightarrow$ decision = 1) \}\\
\end{tabular}
\end{minipage}
%\vspace{2mm}
\caption{Explanations of LORE. (taken from \citet{guidotti2018}. We only use the part of figure 9 of \citet{guidotti2018} that concerns their method, \LORE.) }
\label{fig:lore_example}
\vspace{-3mm}
\end{figure}

A popular method to explain decision on images is the \emph{saliency map}~\cite{adebayo2018, dabkowski2017, fong2017}.
Intuitively, a saliency map higlights the features, \ie the pixels on images, that the model considers to take its decision.
\citet{fong2017} propose an agnostic gradient-based method which considers explanation as meta-predictors.
A meta-predictor is a rule that is used to explain the prediction of the model.
One advantage of using meta-predictors as explanation is that you can measure the ``faithfulness'' of your explanation by computing its prediction error, \ie it represent the number of time that the model and the rule disagree on the prediction.
Then, they claim that a good explanatory rule (\ie meta-predictor) to produce a saliency map rely on the local explanation principle.
Indeed, they want to study the behavior of the model using the neighborhood of the original example.
This neighborhood is given by perturbing the original example; that is why they introduce the notion of ``meaningful pertubations''.
They argue that the explanation would be better, if the perturbations applied to build the neighborhood of the original example mimic \emph{natural image effect}.
Based on meaningful perturbations\footnote{the authors used blur, noise as meaningful perturbation}, the authors define an optimization problem which tries to find the smallest area of the original image that must be perturbed in order to reduce prediction probability of the true class of the original image;
This method then outputs the saliency map which give the explanation of the model's prediction. In table~\ref{tab:XAIMethod}, we named their method Meaningful Perturbation Saliency Map (\MPSM).

\begin{figure}[h]
  \centering
  \resizebox{1\linewidth}{!}{
  \begin{subfigure}[t]{0.24\linewidth}
  \begin{center}
  \includegraphics[width=\linewidth]{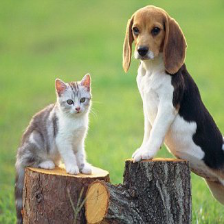}
  \includegraphics[width=\linewidth]{original.png}
  \caption{\tiny Input Image}
  \end{center}
  \end{subfigure}
  \begin{subfigure}[t]{0.24\linewidth}
  \begin{center}
  \includegraphics[width=\linewidth]{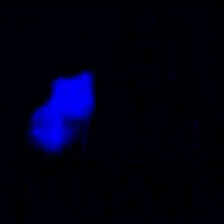}
  \includegraphics[width=\linewidth]{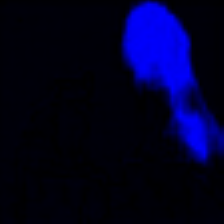}
  \caption{\tiny Generated saliency map}
  \end{center}
  \end{subfigure}
  \begin{subfigure}[t]{0.24\linewidth}
  \begin{center}
  \includegraphics[width=\linewidth]{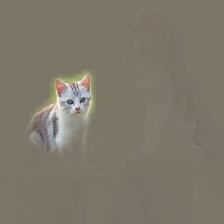}
  \includegraphics[width=\linewidth]{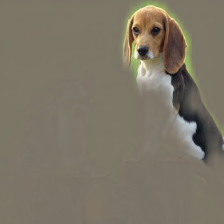}
  \caption{\tiny Image multiplied by the mask}
  \end{center}
  \end{subfigure}
  \begin{subfigure}[t]{0.24\linewidth}
  \begin{center}
  \includegraphics[width=\linewidth]{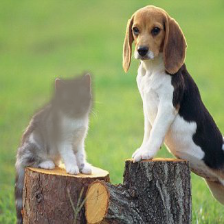}
  \includegraphics[width=\linewidth]{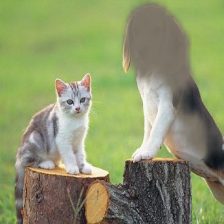}
    \caption{\tiny Image multiplied by inverted mask}
  \end{center}
  \end{subfigure}
  }
  \caption{An example of explanations produced by \citet{dabkowski2017}. The top row shows the explanation for the "Egyptian cat" while the bottom row shows the explanation for the "Beagle". Note that produced explanations can precisely both highlight and remove the selected object from the image.}
  \label{fig:dabkowski2017}
\end{figure}

On the other side, \citet{dabkowski2017} develop a \maskingmodel~that learn how to generate interpretable and accurate saliency map for ML model.
The authors craft a new objective function which ensures the quality of the saliency map: it ensures that the region is necessary to the good classification while its absence leads to low probability to pick the good class and at the same time it penalizes large and not smooth region.
The \maskingmodel~comprises features filter (input) given by a trained network and upsampler layer which upscale the input into the original image dimension in order to provide a mask.
\citet{dabkowski2017} train the \maskingmodel~by directly minimizing its new objective function; the \maskingmodel~outputs saliency map with really accurate salient area (Figure \ref{fig:dabkowski2017}).

\subsubsection{DNN specific}

\citet{adebayo2018} develop a methodology to test the usefulness of a saliency method for explanation.
It is based on two tests: model parameter randomization test and data randomization test.
A saliency method would fail the tests if it shows the same result for the randomized case and the trained case.
A failure points out the independency of the saliency method regarding the architecture parameters or labeled data.
With the methodology proposed by \citet{adebayo2018}, we can test model-agnostic and model-specific methods.

Few years before the works previously presented, \citet{zeiler2014} introduced a method that allow to visualize the internal working of convolutional neural networks (CNN).
Their motivation was to understand how and why CNN work well by studying the hidden layer of CNN.
The principle is to sample the features learned by a neural network, back to the original image dimension.
However, contrary to the method seen until now, the goal of the was not to provide a saliency map that explain the model's predictions but only to visualize the feature that a neural network learned for classifying.
The system develop by \citet{zeiler2014} is known as multi-layered Deconvolutional Network (\deconvnet).
One advantage of this technique is the possibility to start from any layer in order to see at this point the features learned by the neural network.

\begin{figure}
\centering
\includegraphics[width=\textwidth]{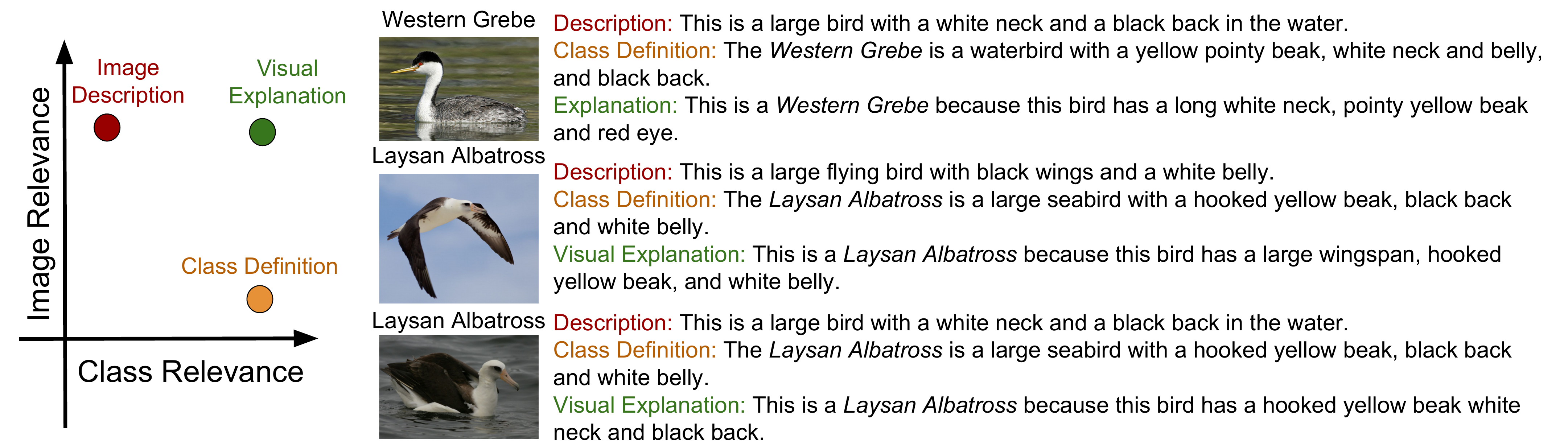}
\vspace{-5pt}
\caption{Visual explanations are both image relevant and class relevant.  In contrast, image descriptions are image relevant, but not necessarily class relevant, and class definitions are class relevant but not necessarily image relevant. (taken from the work of \citet{hendricks2016})}
\vspace{-10pt}
\label{fig:hendricks2016}
\end{figure}

Another way of explaining an image is literally to provide written explanations:
\citet{hendricks2016} propose a \emph{visual explanation} for images\footnote{their method is denoted ``\GVE'' (Generating Visual Explanation) in table~\ref{tab:XAIMethod}.}.
Visual explanations are defined as \emph{class discriminative} and \emph{accurately described}, i.e. the textual explanation highlights elements of the image which are specific to the class.
The authors use VGG, a CNN for image classification, and add to it an LSTM that generates visual explanations; the efficiency of this system relies on the loss function.
Precisely, \citet{hendricks2016} define two losses for their network: one for the accurate description (relevance loss) that handles the probability of word occurrence in the sentence and the other for the class discrimination (discriminative loss) which is based on a reward function.
The authors reach good results on an experiment that deals with bird classification.
Figure~\ref{fig:hendricks2016}, taken from the paper \cite{hendricks2016} shows the difference between visual explanation, image description, and class definition.

\citet{koh_2017} through the use of {\sc influence function} increase in the explanability of DNN.
Unlike the other methods whose the explanability was increased by providing evidences directly on the examples, \citet{koh_2017} to explain the prediction of an example, they provide the training datas that leads to that prediction.
The principle of their method is to compute the impact on the predictions while modifying the training dataset.
For explanability purpose, they study the effect of the removal of training data.
The removed training data which strongly decrease the probability to get the right class of a given example are actually the data that leads to the prediction.
Actually, the training data given as an explanation of a prediction provide insight on the behavior of the model.
Indeed, by looking at the data that influence towards the prediction, it is possible to detect abnormal behavior of the model.

\subsection{Global Explanation}\label{subsec:global_explanation}
As stated earlier, global explanation is the one that could explain the evolution of the model's outputs according to the trends of the inputs.
A popular method is known as ``feature importance''.
The principle of features importance is to find out which input features play a significant role in the decision of the ML algorithm \cite{lundberg2017, shrikumar2017, zhao2019}.

\subsubsection{Model agnostic}
One way of exploit the feature importance principle is to use the Shapley value which comes from cooperative game theory.
The idea is to find the fair share of the gain between players in a cooperative game regarding the involvement of each player in the coalition.
\citet{lundberg2017} develop \SHAP~(SHapley Additive exPlanations) that use the Shapley value in order to provide explanations for the predictions of ML algorithm.
In this context, the features are the players and the prediction is the payout.
Then, \SHAP~compute the contribution of each feature to the prediction.
Since the computation of the Shapley value requires all the possible permutation of the players (features), the problem becomes intractable for inputs with numerous number of features.
However, the authors provide an efficient way to compute an approximation of the Shapley value.

\citet{zhao2019} study the importance of the feature through a causal model, called  Non-Parametric Structural Equation Model (\NPSEM), and using Partial Dependence Plot to visualize the results.
Thanks to this approach they go further than just compute the impact of features on the prediction and manage to catch the tendency of a prediction according to the evolution of features.

\subsubsection{DNN specific}
\citet{shrikumar2017} bring a new way of handling features importance with \deeplift~(Deep Learning Important FeaTures).
The principle is to compare inputs to an input reference and outputs to an output reference.
These references represent a kind of \emph{neutral} behavior of the neural network.
Since these references are considered as baseline, the authors backpropagate the differences between the references and the actual sample through the network by comparing the activation values.
Then, they define a contribution score system that derives the features importance from the differences.
Hence, at the end of the backpropagation, we end up with the importance of each feature for a given prediction.
Intuitively, we would say that the greater the difference is, the more it has an impact on the feature importance.
The efficiency of the algorithm depends on the choice of references that relies on domain-specific knowledge.

\section{Conclusion}
\label{sec:directions}

In this survey we have investigated the compatibility between recent advances in machine learning and established certification procedures used in civil avionic systems.
Machine learning proves to be an interesting approach for developing functionalities that are difficult or impossible to implement using classical programming, such as vision-based navigation (e.g. ATTOL project from Airbus \cite{ATTOL}).
This has the potential to offer meaningful improvements in the area of the system safety, and opens the way to advanced features in autonomous flight systems.
However, the intrinsic characteristics of machine learning based software, such as the use of alternative requirement specification techniques or the difficult traceability of the results, seriously challenge established certification procedures.

As a first step towards certifiable embedded system based on ML, we focused on Adversarial Robustness and Explainability that will eventually support the proof of conformity to regulation requirements.
With Robustness, we address the \emph{innocuity} issue, inherent concept of ML algorithm inducing the necessary management of unintended behaviors of the system.
With Explainability, we address the \emph{trustworthiness} issue in ML algorithms which are still suffering from their black-box aspect.
For both domains, our  review of the state of the art shows promising methods that address those issues.
It is interesting to notice that  most of verification methods (cf. Section~\ref{subsec:verif}) are based on formal methods that have already been used for the demonstration of conformity since a while.
Assuming that verification is the most time and cost consuming part of the development process, formal methods are promising means to handle the robustness aspects because since it provides mathematical evidences, it allows avoiding a huge amount of empirical verification.
Our survey takes a closer look at some particular issues. The ultimate goal of  certifying  embedded systems based on ML, requires to address all the various issues identified in Figure~\ref{fig:certifOverview} and thus demands further research in the other domains.

\bibliographystyle{ACM-Reference-Format}
\bibliography{main}
\end{document}